\documentclass{article}
\usepackage{ijcai22}

\usepackage{times}
\usepackage{soul}
\usepackage{url}
\usepackage[hidelinks]{hyperref}
\usepackage[utf8]{inputenc}
\usepackage[small]{caption}
\usepackage{graphicx}
\usepackage{amsmath}
\usepackage{amsthm}
\usepackage{booktabs}
\usepackage{algorithm}
\usepackage{algorithmic}
\usepackage{array}
\usepackage[numbers]{natbib}

\usepackage{helvet}  % DO NOT CHANGE THIS
\usepackage{courier}  % DO NOT CHANGE THIS
\usepackage{caption} % DO NOT CHANGE THIS AND DO NOT ADD ANY OPTIONS TO IT
\DeclareCaptionStyle{ruled}{labelfont=normalfont,labelsep=colon,strut=off} % DO NOT CHANGE THIS
\usepackage{newfloat}
\usepackage{listings}
\floatstyle{ruled}
\newfloat{listing}{tb}{lst}{}
\floatname{listing}{Listing}

\usepackage{amsmath, amssymb, amsthm}

\title{String-based Molecule Generation via Multi-decoder VAE}

\author{
Kisoo Kwon$^1$
\and
Kuhwan Jung$^1$\and
Junghyun Park$^1$\and
Hwidong Na$^1$\And
Jinwoo Shin$^2$
\affiliations
$^1$Samsung Advanced Institute of Technology, Kyungki, Korea\\
$^2$Korea Advanced Institute of Science and Technology, Daejeon, Korea\\
\emails
\{kisoo.kwon, kuhwan.jung, jhpsy.park, hwidong.na\}@samsung.com,
 jinwoos@kaist.ac.kr}

\def\cL{\mathcal L}

\newcommand{\E}{\mathbb{E}}

\newcommand{\bc}{\begin{center}}
\newcommand{\ec}{\end{center}}
\newcommand{\be}{\begin{equation}}
\newcommand{\ee}{\end{equation}}
\newcommand{\been}{\begin{equation*}}
\newcommand{\eeen}{\end{equation*}}
\newcommand{\ba}{\begin{array}}
\newcommand{\ea}{\end{array}}
\newcommand{\bean}{\setlength\arraycolsep{2pt}\begin{eqnarray*}}
\newcommand{\eean}{\end{eqnarray*}}
\newcommand{\bea}{\setlength\arraycolsep{2pt}\begin{eqnarray}}
\newcommand{\eea}{\end{eqnarray}}
\newcommand{\ben}{\begin{enumerate}}
\newcommand{\een}{\end{enumerate}}
\newcommand{\bed}{\begin{itemize}}
\newcommand{\eed}{\end{itemize}}

\begin{document}

\maketitle

\begin{abstract}
In this paper, we investigate the problem of string-based molecular generation via 
variational autoencoders (VAEs) that have served a popular generative approach for various tasks in artificial intelligence. We propose a simple, yet effective idea to improve the performance of VAE for the task.
%, particularly focusing on the case when decoders are auto-regressive models. 
Our main idea is to maintain multiple decoders while sharing a single encoder, i.e., it is a type of ensemble techniques. Here, we first found that training each decoder independently may not be effective as the bias of the ensemble decoder increases severely under its auto-regressive inference. 
To maintain both small bias and variance of the ensemble model, our proposed technique is two-fold:
(a) a different latent variable is sampled for each decoder (from estimated mean and variance offered by the shared encoder) %given a molecule 
to encourage diverse characteristics of decoders
and (b)
%we suggest to use
a collaborative loss is used during training to control the aggregated quality of decoders using 
different latent variables. % for each decoder during training.
In our experiments, 
%We demonstrate the superiority of the proposed scheme under the string-based molecular generation task: in particular, 
the proposed VAE model particularly performs well for generating a sample from out-of-domain distribution.
\end{abstract}

\section{Introduction}

%Deep learning techniques have shown the remarkable performance not only for the standard prediction task, e.g., classification and regression, in artificial intelligence, but also for generation tasks, e.g., 
%produce photo-realistic images indistinguishable to real images \cite{Karras2020training,Hudson2021gerative}. Deep generative models have been also actively applied to applications such as story generation \cite{Fan18Hier} and text-to-image translation \cite{Li16deep,Gu18dialogwae}.

For the material discovery via molecular generation, several approaches based on machine learning are actively adopting
%In this paper, we consider the task of material discovery via molecular generation
\cite{sanchez2018inverse,shi2020graph,jin2018junction,gomez2018automatic,dai2018syntax}. Developing a novel molecule with desired properties is challenging because it costs huge amount of time and budget for laboratory experiment. We aim to provide an effective deep learning model to generate novel molecules for reducing the cost. Instead of human experts that design novel molecules using the domain knowledge, such a machine learning model can do a similar job by generating candidate molecules based on existing data annotated with physical properties. In order to enforce generated candidates to have the desired properties, target properties are given in addition to the molecule itself during training.

For generating string representation of novel molecules, we investigate variational autoencoders (VAEs) \cite{kingma2014auto} with auto-regressive decoders, which is of popular choice as a deep generative model for such a generation task.
VAE minimizes Kullback-Leibler divergence (KLD) of variational approximation from the true posterior by maximizing evidence lower bound (ELBO). 
Although extensive research efforts have been made on improving vanilla VAE models, they typically focus on traditional AI domains such as images \cite{Pu16Advances}, languages \cite{dai2018syntax} and speech \cite{Wei17Unsuper}; it is not clear whether such complex techniques are useful for our task %of molecular generation. 
to find novel molecules
presumably following the out-of-distribution (OOD) domain.
%Indeed, 

Instead, we aim for providing an orthogonal simple, yet effective idea. To this end, we use `ensemble' techniques that are arguably the most trustworthy techniques for boosting the performance of machine learning models. They have played a critical role in the
machine learning community to obtain better predictive
performance than what could be obtained from any of
the constituent learning models alone, e.g., Bayesian
model/parameter averaging \cite{Domingos00Bayesian}, boosting
\cite{Freund99short} and bagging \cite{Breiman96Bagging}.
However, 
%a vanilla VAE model conditioned on target properties produces good candidates, a single model may not be enough as the search space is exponentially large. Ensemble techniques utilizing multiple models have successfully proved ``two heads are better than one'' in various tasks \cite{Ren2016ensemble} and domains \cite{Cao2020ensemble}. An ensemble model generally obtains more stable outcomes via collaboration of less accurate models. 
under the VAE framework, researchers draw less attention to ensemble techniques
as it is not straightforward how to aggregate predictions of 
multiple encoders and decoders of VAE for improving them.

%than other research fields. A natural question arises from this perspective, ``Can we make a VAE better using ensemble techniques?''

%To address the question, 
In this paper, we design an ensemble version of VAE by utilizing multiple decoders in VAE, while sharing the encoder. Specifically, our proposed idea ensembles logits from decoders in an auto-regressive generation process. 
Here, we found that training each decoder independently is not effective as the bias of the ensemble model increases under its auto-regressive inference.\footnote{ 
%Here, we found that training each decoder independently is not effective as the bias of the ensemble model increases severely under its auto-regressive inference. 
For example, in our experiment, multiple decoders trained independently showed a higher reconstruction loss than vanilla VAE with the similar model size.}
To maintain both small bias and variance of the ensemble model, each decoder is trained not only individually, but also in a way to collaborate with the others. 
Furthermore, to encourage diverse characteristics of decoders, we also propose to sample a different latent variable for each decoder (from estimated mean and variance offered by the shared encoder) given a molecule during training (see Figure \ref{fig:MD} for the illustration of our proposed scheme).

We conducted experiments on a publicly available dataset on the molecular generation task for demonstrating the superiority of the proposed multi-decoder VAE model (MD-VAE). MD-VAE achieved lower training loss and higher reconstruction accuracy than baselines including VAE with k-annealing and Controllable VAE \cite{bowman2016generating,Shao2020control}.
As aforementioned,
%Note that we aim to find novel molecules presumably following the out-of-distribution domain (OOD), and 
the efficiency of generation for OOD-domain is more important than that for in-domain. We emphasis that MD-VAE achieved 31.4\% higher relative generative efficiency than Controllable VAE (ControlVAE) for this perspective (ControlVAE: $36.3\% \rightarrow$ MD-VAE: $47.7\%$). 
This is remarkable as
VAEs are often reported to be poor for generating OOD-domain samples \cite{montero2020role}.
Although we apply MD-VAE to the molecular generation task, 
we believe that it is of interest for a broader range of domains, e.g image generation with desired styles or text generation with desired sentiments.

\section{Related Work}
\subsection{Ensemble method} 
%,Rokach10Ensemble
Ensemble is arguably the most trustworthy technique or concept to improve the performance of a given machine learning model \cite{Opitz99Popular,Wang20ensemble}. The ensemble method gives room to appropriately control the trade-off between bias and variance of the model.
The effect of ensemble is largely associated with the expertise of individual models. That is, a diversity of models is a one of essential factor for a success of ensemble \cite{Kuncheva03measures}.
Due to the reason, many ensemble methods seek to promote diversity among the models they combine \cite{Brown05Diversity}.
As a standard choice, multiple predictive models trained independently are averaged in order to obtain a ensemble effect.
Recently, several attempts have been made to improve the performance by directly learning the (collaborative) loss of ensemble models \cite{wang2021long}.
While the preceding methods on this line consider the standard classification tasks, 
our proposed method targets a more complex task of auto-regressive generation that  
each token is generated by ensembles of each decoder's logit.

\subsection{Molecular design} %(molecular generation)} 
Traditional molecular design using domain knowledge requires simulation techniques due to time and cost issues. High throughput computational screening, a brute-force simulation of a large number of molecules, has been conducted \cite{bleicher2003hit}, but recently, a research direction using deep learning is being actively explored, namely inverse design. It is a study on a model that generates molecules that match the given physical properties. Studies applying generative models such as VAEs are in progress in this field, and they can be roughly divided into two directions according to the molecular expression. The first one expresses molecules as sequences (or strings) such as SMILES \cite{weininger1989smiles} while the other one does molecules as graphs \cite{jin2018junction}.

\begin{figure*}[ht]
\vspace{-0.15in}
  \centering
  \begin{tabular}{cc}
     \multicolumn{2}{c}{\includegraphics[width=0.8\textwidth]{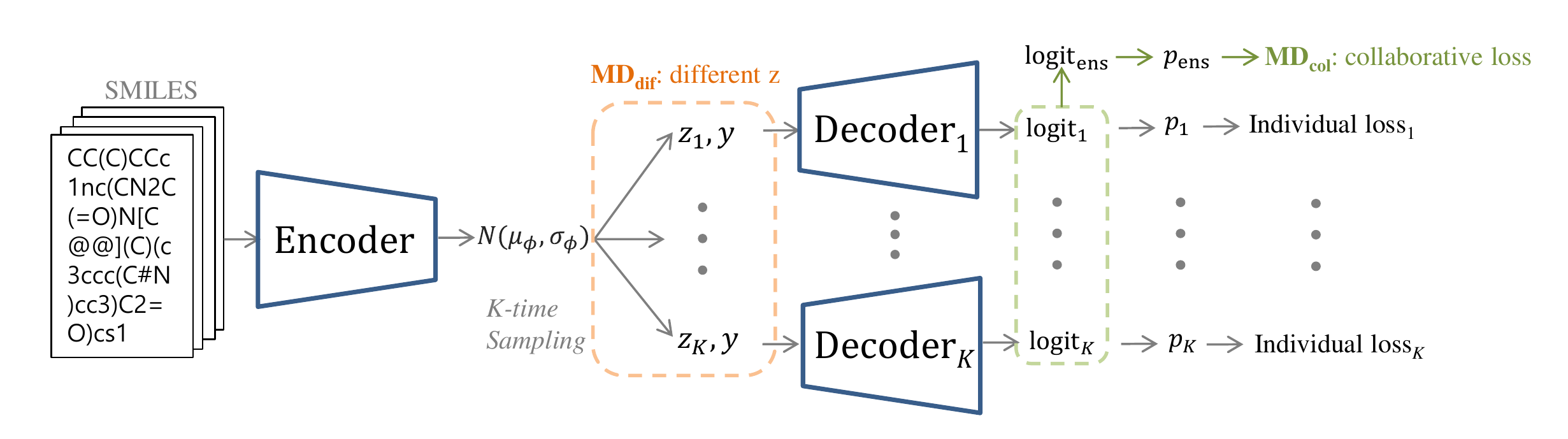}}\\
     \multicolumn{2}{c}{(a) Structure of MD-VAE}\\
     \includegraphics[width=0.35\textwidth]{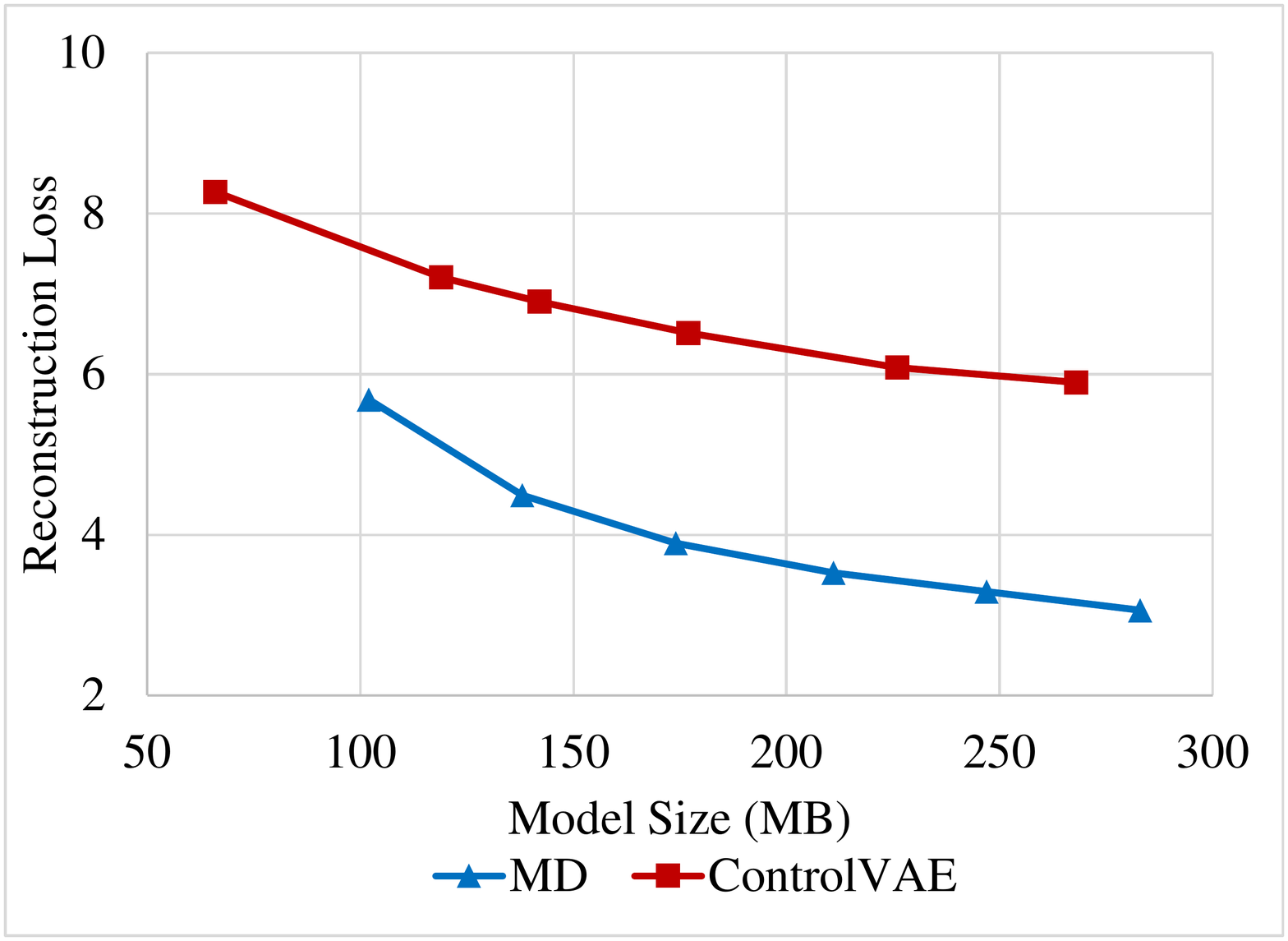} & 
     \includegraphics[width=0.35\textwidth]{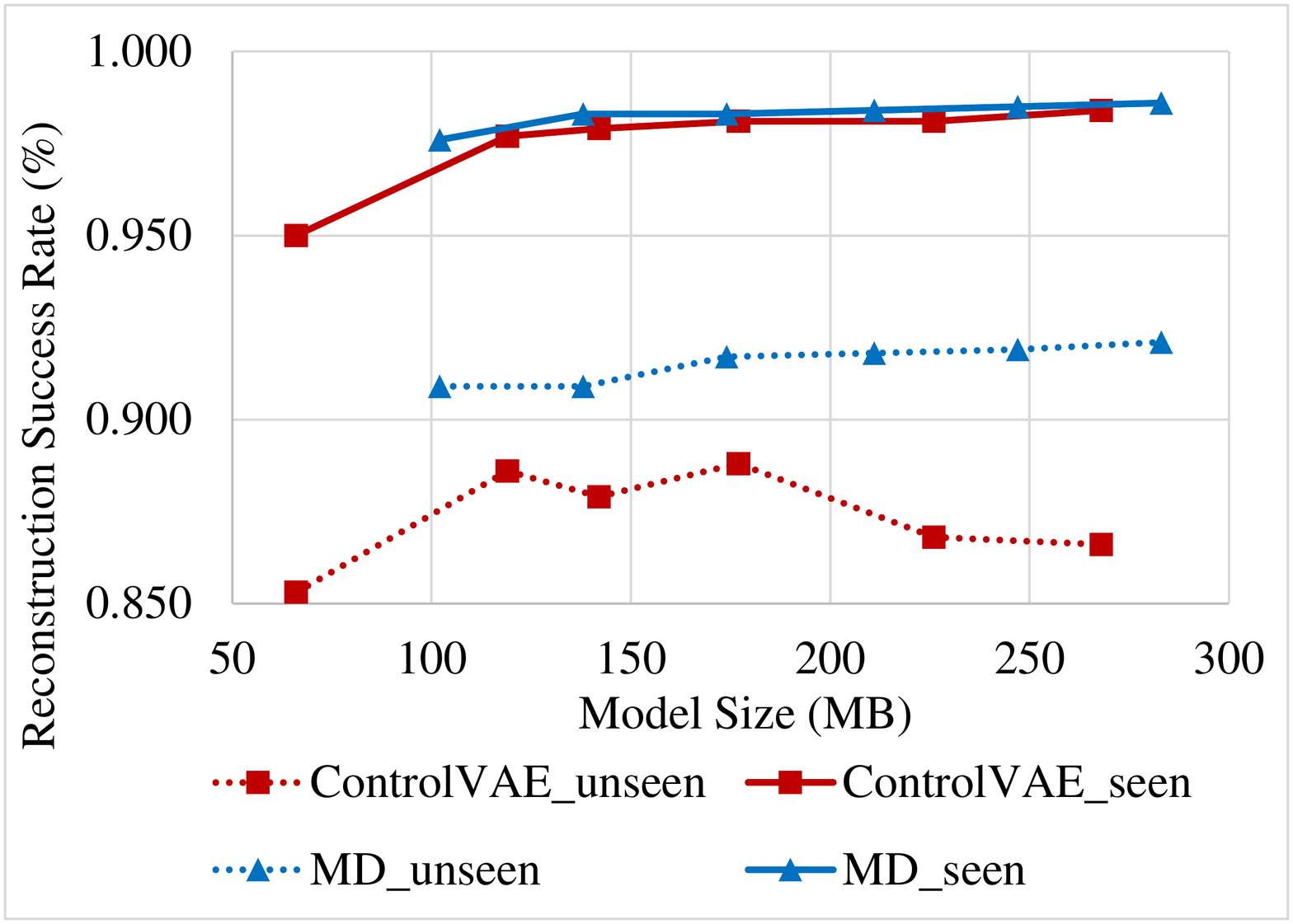}\\
     (b) Reconstruction Loss & 
     (c) Reconstruction Success Rate (\%)\\
  \end{tabular}
  \caption{
  (a) Structure of MD-VAE. (b) indicates a reconstruction loss according to the number of decoders (from 2 to 7). The $x$-axis denotes the model size (mb). The encoder and each decoder size of MD-VAE (\textit{MD}$_\text{dif,col}$) is the same. As the number of decoders increases, the model size of MD-VAE also be increased. MD-VAE outperformed ControlVAE in all model size. 
  (c) The reconstruction success rate was evaluated under a teacher-forcing. The performance gap between MD-VAE and ControlVAE is remarkably large in the unseen case. 
  }
  \label{fig:MD}
  \vspace{-0.15in}
\end{figure*}

\section{Conditional and Controllable VAE}

%  [1. cVAE 소개, transformer 기반] 

\label{sec:vae}
The conditional VAE (cVAE) \cite{kang2018conditional,kingma2014semi} is designed to generate data given certain conditions such as classes or labels.
In the cVAE, SMILES of molecule $x$ is assumed to be generated from $p_\theta(x|y,z)$ conditioned on target molecular properties $y$ and latent variable $z$.
The prior distribution of $z$ is assumed to be Gaussian distribution, i.e. $p(z)=\mathcal{N}(z|0, I)$.
We use variational inference to approximate the posterior distribution of $z$ given $x$ and $y$ by
\begin{equation}
\label{eq:encoder}
q_\phi(z|x,y)=\mathcal{N}(z|\mu_\phi(x,y),\text{diag}(\sigma_\phi(x,y))).
\end{equation}

From the perspective of the auto-encoder, $q_\phi(z|x,y)$ and $p_\theta(x|y,z)$ are called as an encoder and a decoder, respectively.
To deal with string data, a structure such as Gated Recurrent Unit (GRU) \cite{Cho14learning} or Long Short-Term Memory (LSTM) \cite{Sak14long} is usually used. In this paper, since transformer \cite{Vaswani2017attention} shows a better performance in cVAE than GRU and LSTM, transformer encoders are used for $\mu_\phi(x,y)$ and $\sigma_\phi(x,y)$ and a transformer decoder is used for $p_\theta(x|y,z)$.

The objective of the cVAE is to maximize ELBO, which is a lower bound of the marginal log-likelihood:
\bea
\label{eq:elbo}
&&\log p_\theta(x|y)\\ \nonumber
&&\geq\E_{q_\phi(z|x,y)}[\log p_\theta(x|y,z)]-\text{KLD}(q_\phi(\cdot|x,y)||p(\cdot)).
\eea

We define $\mathcal{L}_{\rm recon}=-\E_{q_\phi(z|x,y)}[ \log p_\theta(x|y,z)]$ because it can be regarded as a reconstruction loss.
$\mathcal{L}_{\rm reg} = \text{KLD}(q_\phi(\cdot|x,y)||p(\cdot))$ is acting like a regularizing term.
In summary, parameters $\theta$ and $\phi$ are jointly optimized to minimize $\mathcal{L}_{\rm total} = \mathcal{L}_{\rm recon} + \mathcal{L}_{\rm reg}$.
Given a target molecular properties $y$, a new molecule $x$ having this property is generated in the following way:
\begin{equation}
\label{eq:infer}
z\sim p(z),~x\sim p_\theta(x|y,z).
\end{equation}

Various ideas have been proposed to avoid the posterior vanishing problem in VAE models. VAE with k-annealing \cite{bowman2016generating} and its variants have been shown that it is possible to prevent the posterior vanishing problem using an appropriate scale $\beta$ of the regularizing term, i.e. $\mathcal{L}_\text{total} = \mathcal{L}_\text{recon} + \beta\mathcal{L}_\text{reg}$. Since it is difficult to accurately compare the performance among generative models by exploring the degree of convergence of $\mathcal{L}_{\rm reg}$ with a hyper-parameter, it is practically useful compared to learning a model with $\beta$ as a hyper-parameter. ControlVAE \cite{Shao2020control} further enhanced the idea by automatically tuning $\beta$ to converge $\mathcal{L}_{\rm reg}$ to a specific value based on automatic control theory. This idea does not necessarily lead to convergence to the best $\mathcal{L}_{\rm reg}$, but it is possible to model the converging a desired $\mathcal{L}_{\rm reg}$. 
$\beta$ is updated to have a target $\mathcal{L}_{\rm reg}$ at every training steps $t$:
%The difference of the target $\mathcal{L}_{\rm reg}$ $v^*$ and the sampled $\mathcal{L}_{\rm reg}$ $v(t)$, i.e. $e(t) = v^* - v(t)$, is the feedback at the training step $t$, and $\beta(t) = \frac{K_p}{1 + \exp(e(t))} - K_i \sum\nolimits_{j=0}^{t} e(j) + \beta_{\min}$ is automatically adjusted by a controller with two hyper-parameters $K_p$ and $K_i$:
\begin{equation}
\label{eq:control-vae}
\mathcal{L}_{\rm total} = \mathcal{L}_{\rm recon} + \beta(t) \mathcal{L}_{\rm reg}.
\end{equation}

%We applied ControlVAE to both the baseline and proposed method experiments, and conducted experiments on various KLD hyper-parameters, and compared the best model performance under the same KLD conditions. Hence we focus on the reconstruction loss $\mathcal{L}_\text{recon}$ in the rest of this article.

\section{Multi-decoder VAE}

In this paper, we propose a neural architecture for VAE, which we name MD-VAE.
Our main idea is to use multiple decoders while sharing a single encoder, as illustrated in Figure~\ref{fig:MD}. 
In particular, we consider an auto-regressive model (e.g., recurrent neural network or transformer)
as each decoder architecture. Then, to aggregate outputs of multiple decoders,
the decoder's logit values are averaged to predict the next token in an auto-regressive manner.\footnote{We tested two different schemes in the ensemble computation: pre-softmax (logit-level) and post-softmax (probability-level) interpolation. In our preliminary experiments, we validated that the difference between two schemes in various metrics we used are negligible, but pre-softmax converges faster than post-softmax interpolation. Hence we conducted the rest of experiments using the pre-softmax scheme, i.e., we average the logits instead of the softmax probabilities.}  Here, each decoder has its own separate parameters, and produces a different logit value. One may expect that such ensemble of different logits provides more robust prediction than that from an individual decoder. However, to the best of our knowledge, such ensemble version of VAE has not been explored in literature. 

This is because
a naive ensemble of independently trained decoders (while sharing a single encoder) 
may increase the model bias (although it may decrease the model variance), as
it can boost up significantly under VAEs using auto-regressive decoders.
To mitigate such an negative effect,
we propose to optimize the following additional collaborative loss to train our model: 
\begin{equation}
    \mathcal{L}_{\rm col} = - \E_z\left[\log \frac{1}{K} \sum\nolimits_k p_{\theta_k} (x |y, z)\right]
    \label{eq:loss1}
\end{equation}

Herein $K$ indicates the number of decoders of MD-VAE. We remark that such a collaborative loss may not be effective for ensemble of non-auto-regressive models, e.g., the standard classification or regression model, as it increases the model variance. However, in our case using auto-regressive models, it is effective as reducing the model bias is more crucial than doing the model variance. Nevertheless, even for our model,
to reduce the model variance, we suggest to sample
a different latent variable from the shared encoder for each decoder during training.
It encourage for decoders to produce diverse outputs, and hence reduces the variance of the ensemble model.
Specifically, each latent variable is sampled from the approximation of the posterior distribution of $z$ given $x$ and $y$, and there are $K$ sampled latent variables $z_1, z_2, \ldots, z_K$ where
\begin{equation}
 z_k \sim \mathcal{N}(z_k|\mu_\phi(x,y),\text{diag}(\sigma_\phi(x,y))).
\end{equation}

In summary, we integrate two ideas to train the proposed MD-VAE model. 
The collaborative loss promotes small bias of the ensemble prediction over multiple decoders, while 
sampling different latent variables for decoders does small variance of the ensemble model.
Namely, we consider the following additional loss to train decoders in MD-VAE:
\begin{equation}
    \mathcal{L}_\text{dif,col} = - \E_{z_1,\ldots,z_K}\left[\log \frac{1}{K} \sum\nolimits_k p_{\theta_k} (x |y, z_k)\right].
    \label{eq:loss3}
\end{equation}

The only difference in Eq.~\eqref{eq:loss1} and Eq.~\eqref{eq:loss3} is to use the latent variable $z$ or $z_k$. 

The total reconstruction loss is a linear interpolation between collaborative and individual loss functions:
\begin{equation}
\mathcal{L}_\text{recon}^\text{MD} = \alpha \mathcal{L}_\text{dif,col} - \frac{(1-\alpha)}{K} \sum_{k=1}^{K} \log p_{\theta_k} (x | y, z_k).
\label{eq:loss-recon}
\end{equation}
Eq.~\eqref{eq:loss-recon} is minimized together with the regularizing term $\mathcal{L}_\text{reg}$ in Eq.~\eqref{eq:control-vae}:

\begin{equation}
\label{eq:final}
\cL_\text{total}^\text{MD} = \cL_\text{recon}^\text{MD} + \beta(t) \cL_{\rm reg}.
\end{equation}

The two ideas mentioned above (using multiple latent variables and using collaborative loss) are also applicable to vanilla single-decoder VAE (SD-VAE).
In this case, SD-VAE can not utilize an ensemble scheme at an inference phase. 
%More specifically, SD-VAE using $K$ latent variables is called SD$_\text{dif}$-VAE (simply \textbf{SD}$_\text{dif}$), and the one additionally using collaborative loss is called SD$_\text{dif,col}$-VAE (simply \textbf{SD}$_\text{dif,col}$).
%We also use similar abbreviations (\textbf{MD}$_\text{col}$, \textbf{MD}$_\text{dif}$ and \textbf{MD}$_\text{dif,col}$) for MD-VAE.

\section{Experiments}
\label{sec:experiment}

\begin{table}
    \centering
    \caption{The model size, reconstruction loss, regularizing loss and KLD of decoders (ZINC250K dataset). Each model's regularizing loss is controlled by 15 without \textit{Base}. (Each model trained 3-time, and each figure was the average of 3 results.)}
    \label{tab:loss}
    \begin{tabular}{|l|c|c|c|c|}
    \hline
        Model & Size & $\mathcal{L}_{\rm recon}$ & $\mathcal{L}_{\rm reg}$ & KLD \\  \hline \hline
        \textit{Base} & 142MB & 17.276 & 0.000 & -  \\ \hline
        \textit{ControlVAE} & 142MB & 6.851 & 15.168 & - \\ \hline
        $\textit{SD}_\text{dif,col}$ & 142MB & \textbf{6.657} & 15.189 & - \\ \hline
        \textit{MD} & 138MB & 7.001  & 15.207 & 0.2179 \\ \hline
        $\textit{MD}_\text{col}$ & 138MB & \textbf{5.508} & 14.937 & 0.5582 \\ \hline
        $\textit{MD}_\text{dif}$ & 138MB & \textbf{6.555} & 15.145 & 0.2723\\ \hline
        $\textit{MD}_\text{dif,col}$ & 138MB & \textbf{4.482} & 15.068 & 0.4384\\ \hline
    \end{tabular}
\end{table}

\begin{figure}
  \centering
  \includegraphics[width=0.2\textwidth]{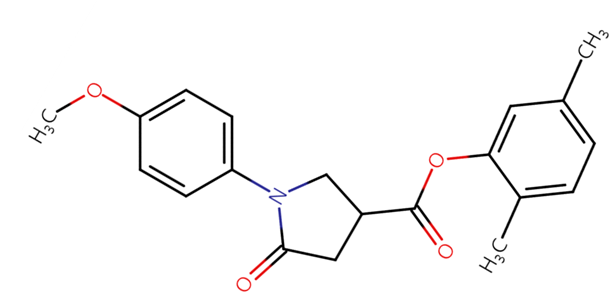}
  \caption{An example of SMILES in ZINC: COc1ccc(N2CC(C(=O)Oc3cc(C)ccc3C)CC2=O)cc1.}
  \label{fig:SMILES}
\end{figure}

\subsection{Experiment Setup}
\label{subsec:setup}

ZINC \cite{sterling2015zinc} is a database comprising information on various drug-like molecules. 
ZINC contains 3D structural information of molecules and molecular physical properties such as molecular weight (molWt), partition coefficient (LogP), and quantitative estimation of drug-likeness (QED). 
Figure~\ref{fig:SMILES} shows SMILES of a molecule in ZINC, and we used two subsets of ZINC: ZINC250K \cite{kusner2017grammar} and ZINC310K \cite{yan2019re,kang2018conditional}. The vocabulary for SMILES contains 39 different symbols \{1, 2, 3, 4, 5, 6, 7, 8, 9, +, -, =, \#, (, ), [, ], H, B, C, N, O, F, Si, P, S, Cl, Br, Sn, I, c, n, o, p, s, ${\backslash}$, /, @, @@\}. The minimum, median, and maximum lengths of a SMILES string in ZINC250K are 9, 44, and 120, respectively (In case of ZINC310K, 8, 42, and 86).
%The distribution of target properties in ZINC250K and ZINC310K is shown in Figure~\ref{fig:distrib}. 
The average values of molwt, LogP, and QED are about 330, 2.457, and 0.7318 in ZINK250K, respectively.
The three quantities are 313, 1.9029, 0.7527 in ZINK310K, respectively.

\begin{table}
    \centering
    \caption{The reconstruction success rate for seen (ZINC250K) and unseen (ZINC310K) dataset. The number in boldface means that it is better than the \textit{ControlVAE}. (Each figure was the average of 3 results.}
    \label{tab:recon}
    \begin{tabular}{|l|c|c|c|c|}
    \hline
        Model & \bf{Unseen} & Seen & Average \\ \hline \hline
        \textit{Base} & 0.783 & 0.906 & 0.845  \\ \hline
        \textit{ControlVAE} & 0.880 & 0.979 & 0.930 \\ \hline
        $\textit{SD}_\text{dif,col}$ & 0.880 & \bf{0.989} & \bf{0.935} \\ \hline
        $\textit{MD}$ & \bf{0.898} & 0.974 & \bf{0.936}  \\ \hline
        $\textit{MD}_\text{col}$ & \bf{0.891} & \bf{0.978} & \bf{0.934} \\ \hline
        $\textit{MD}_\text{dif}$ & \bf{0.902} & 0.975 & \bf{0.938} \\ \hline
        $\textit{MD}_\text{dif,col}$ & \bf{0.909} & \bf{0.983} & \bf{0.946} \\ \hline
    \end{tabular}
\end{table}

%\begin{figure}
  %\centering
    %\begin{tabular}{c}
%     \includegraphics[width=0.45 \textwidth]{dist1-1.PNG}\\
	%(a) ZINC250K\\
     %\includegraphics[width=0.45 \textwidth]{dist1-2.PNG}\\
	%(b) ZINC310K\\
%  \end{tabular}
%  \caption{Distributions of target properties in (a) ZINC250K and (b) ZINC310K. ZINC250K was used in the
%  training phase.}
%  \label{fig:distrib}
%\end{figure}

\begin{table*}[ht]
    \centering
    \caption{The molecular generative efficiency (Probability that all three conditions, validity, uniqueness and novelty, are satisfied): For generative efficiency, the condition values of conditional VAE were divided into two cases. For in-domain case, we chose three condition values based on the property distribution at each property. For y-extrapolation (OOD) case, two condition values ($\mu \pm 4\sigma$) were used. At each condition value, 2,000-times generation was tried. The number in boldface means that it is better than \textit{ControlVAE}.}
    \label{tab:eff}
    {\small
    \begin{tabular}{|l|c|c|c|c|c|c|c|c|}
    \hline
         & \multicolumn{4}{c|}{In-domain} & \multicolumn{4}{c|}{Out-of-distribution domain} \\ \cline{2-9}
         & molWt & logP & QED & Average & molWt & logP & QED & Average \\ \hline \hline
        \textit{Base} & 0.224 & 0.308 & 0.329 & 0.287 & 0.100 & 0.069 & 0.110 & 0.0929 \\ \hline
        \textit{ControlVAE} & 0.910 & 0.916 & 0.913 & 0.913 & 0.377 & 0.311 & 0.403 & 0.363\\ \hline 
        $\textit{SD}_\text{dif,col}$ & 0.903  & 0.907  & 0.911 & 0.907 & 0.307 & 0.278 & 0.398 & 0.328 \\ \hline
        \textit{MD} & \textbf{0.946} & \textbf{0.944} & \textbf{0.946} & \textbf{0.945} & \textbf{0.489} & \textbf{0.415}  & \textbf{0.472} & \textbf{0.458}  \\ \hline
        $\textit{MD}_\text{col}$ & \textbf{0.915}  & \textbf{0.922} & 0.909 & \textbf{0.915} & \textbf{0.431} & \textbf{0.398} & \textbf{0.414} & \textbf{0.414}\\ \hline
        $\textit{MD}_\text{dif}$ & \textbf{0.928} &\textbf{ 0.936 }& \textbf{0.938} & \textbf{0.934} & \textbf{0.457} & \textbf{0.427}  & \textbf{0.468}  & \textbf{0.451}  \\ \hline
        $\textit{MD}_\text{dif,col}$ & \textbf{0.935} & \textbf{0.933} & \textbf{0.932} & \textbf{0.933} & \textbf{0.457} & \textbf{0.539} & \textbf{0.436} & \textbf{0.477}  \\ \hline
    \end{tabular}
    }
\end{table*}

Each encoder and decoder of the VAE model consists of three layers with self-attention like transformer, and the dimension of the latent variable is set to 100, and the dimension of the properties (conditions) is 3. 
%A 103-dimensional vector in which a 100-dimensional latent variable and a 3-dimensional property are concatenated is inputted to the decoder. 
We used the Adam optimizer \cite{Ruder16overview} with $\beta_1=0.9$, $\beta_2=0.999$, and $\epsilon=10^{-6}$, and the initial learning rate was 0.001. Each model was trained during 100 epochs\footnote{In our experimental results, the single decoder based VAE models which were trained until 1,000 epochs showed a worse reconstruction success rate for unseen dataset (ZINC310K) than the models which were trained until 100 epochs. Therefore, the average reconstruction success rate of seen and unseen dataset of 100 epochs also better than those of 1,000 epochs. It also showed a better molecular generative efficiency when each model trained until 100 epochs.}, and batch size was 128.
%In the case of a molecular generation, since the performance for OOD or unseen case is more important than in-domain case, we fixed the training epoch to 100.
%During the training, we normalized each property value to have a mean 0 and standard deviation 1. 
%Each model was trained during 1000 epochs, and it showed stable training loss with several random initial weights. For example, the maximum relative difference between three runs of the baseline model was less than 0.5\%, and other models showed the similar trends.

We compared five methods and its detailed information is as follows:

\begin{footnotesize}
$\bullet$ \textit{Base}: vanilla VAE with k-annealing \cite{bowman2016generating}

$\bullet$ \textit{ControlVAE}: controllabe VAE \cite{Shao2020control}. Vanilla VAE suffered a posterior collapsing in molecular domain even if it adopted k-annealing. Therefore, small $\beta$ for $\cL_\text{reg}$ helps to reduce the posterior collapsing, also it enhances a molecular generative performance \cite{yan2019re}. According to the our experimental results, when $\cL_\text{reg}$ was controlled as about 15, \textit{ControlVAE} showed a proper performance. 

$\bullet$ \textit{SD}$_\text{dif,col}$: \textit{ControlVAE} with 3-time $z$ sampling and collaborative loss. However, it can not utilize a diverse speciality of multi-decoder.

$\bullet$ \textit{MD}: \textit{ControlVAE} with multi-decoder with only the individual loss and the same sampled $z$.

$\bullet$ \textit{MD}$_\text{col}$: \textit{ControlVAE} with multi-decoder and collaborative loss. 

$\bullet$ \textit{MD}$_\text{dif}$: \textit{ControlVAE} with multi-decoder and multi-$z$ sampling. Each decoder is trained by different $z$ from the same input data.

$\bullet$ \textit{MD}$_\text{dif,col}$: \textit{ControlVAE} with multi-decoder, collaborative loss and multi-$z$ sampling.
\end{footnotesize}

\subsection{Evaluation: Training Phase}
\label{subsec:loss4}
ZINC250K was used for the training data, and the reconstruction loss of each model was measured for the evaluation metric.
In the cases of single decoder based VAE (\textit{Base}, \textit{ControlVAE}, \textit{SD}$_\text{dif,col}$), the reconstruction loss is the sum of the cross-entropy for each token between the decoder's output and the true label. 
For \textit{MD}s (\textit{MD}, \textit{MD}$_\text{col}$, \textit{MD}$_\text{dif}$ and \textit{MD}$_\text{dif,col}$), an ensemble of decoders' outputs is used as an output for the reconstruction loss.

We fixed the number of decoders in \textit{MD}s as 3, because it can utilize advantages of the proposed method sufficiently. Of course, for a reasonable comparison, the model size of \textit{MD}s was adjusted to be the same as the size of \textit{SD}s. (Each size of decoder of \textit{MD}s is the same as the size of encoder.)

The size and reconstruction loss of each model is shown in Table~\ref{tab:loss}. 
Although we adopted k-annealing to \textit{Base}, its training results seemed to be not proper. \textit{ControlVAE} showed a far better performance than \textit{Base}, so we decided to compare our proposed methods with \textit{ControlVAE}.
All MD-VAE models showed smaller reconstruction loss than \textit{ControlVAE} except \textit{MD}. 
%In the single-decoder case, our proposed method \textbf{SD}$_\text{dif,col}$ also achieved smaller reconstruction loss than \textbf{ContVAE}. 
Through these comparisons, it can be seen that the proposal methods are effective to reduce the reconstruction error during training. 
We also measured KLD between the 3 decoders of \textit{MD}s, and it was confirmed that KLD of \textit{MD}$_\text{dif,col}$ (0.4384) was greater than that of \textit{MD} (0.2178).
This means that using different latent variables and collaborative loss for decoders in MD-VAE would result in decoders having more different features, which import to enhance the generalization capacity.

\subsection{Evaluation: Reconstruction Success Rate}
\label{subsec:recon_eval}

It is important how accurately restore the molecule $x$ from $\mu_\phi(x,y)$ through the decoder.
This can be seen that $z-$space is appropriately expressing molecule space of not only the training dataset but also an unseen dataset. In order to generate a new molecule, the reconstruction success rate on unseen data is crucial point. 
To verify this, we checked that molecules were properly restored at token-level.
This evaluation was performed in a teacher-forcing situation without a reparameterization trick.
We calculated reconstruction success rates for the training (ZINC250K) and unseen (ZINC310K) dataset in Table~\ref{tab:recon}.
For this experiment, duplicate samples were removed in ZINC310K.
Every \textit{MD}s outperformed \textit{Base} and \textit{ControlVAE} in the seen and unseen cases. In particular, \textit{MD}s showed a bigger improvement when OOD conditions were used. 
This tendency grows with longer training epoch (When each model was trained until 1,000 epochs, \textit{ControlVAE} and \textit{MD}$_\text{dif,col}$ showed 0.846 and 0.894 in terms of reconstruction success rate for the unseen case, respectively.).
%In a nutshell, the proposed \textit{MD}s averagely showed a good reconstruction performance, and it can be proper to generate new molecules.

\begin{table*}[ht]
    \centering
    \caption{The conditional satisfaction in terms of MAE between a condition and simulation value of the top 1 molecules generated by each model (the lower the better): 
    For a conditional molecular generation, the quality of generated molecules is one of the essential factor.
    For each condition and property, the best generated molecules are compared. For each case, molecular generation was tried 2,000 times by each model. ($\text{QED}=1.2861$ was an improper condition because it was a physically absent region. The maximum QED from RDKIT is 0.948.)}
    \label{tab:satis}
    {\small
    \begin{tabular}{|l|c|c|c|c|c|c|c|c|c|}
    \hline
         Top1 MAE &\multicolumn{3}{c|}{In-domain} & \multicolumn{6}{c|}{Out-of-distribution domain} \\ \hline
         Property & molWt & LogP & QED & \multicolumn{2}{c|}{molWt} & \multicolumn{2}{c|}{LogP} & \multicolumn{2}{c|}{QED} \\ \hline
         Condition & {\scriptsize 434, 330, 230} & {\scriptsize 4.816, 2.457, 0.098} & {\scriptsize 0.9598, 0.7318, 0.5038} & 580 & 84 & 8.194 & -3.281 & 1.2861 & 0.1778 \\ \hline \hline
        \textit{Base} & 0.1520 & 0.0008 & 0.0041 & 0.0810 & 0.0780 & 0.0013 & - & - & 0.0006 \\ \hline
        \textit{ControlVAE} & 0.1177 & 0.0005 & 0.0041 & 0.0800 & 0.0740 & 0.0005 & 1.3598 & 0.5771 & 0.0008 \\ \hline
        $\textit{SD}_\text{dif,col}$ & \bf{0.0850} & \bf{0.0002} & 0.0044 & \bf{0.0370} & 0.0780 & \textbf{0.0001} & - & 0.6720 & \textbf{0.0002} \\ \hline
        \textit{MD} & \bf{0.0940} & \bf{0.0003} & \textbf{0.0040} & 0.1740 & 0.0980 & \bf{0.0003} & \textbf{0.0204} & \textbf{0.4405} & \bf{0.0015} \\ \hline
        $\textit{MD}_\text{col}$ & \bf{0.0497} & \bf{0.0013} & 0.0042 &\textbf{0.0760} & \bf{0.0740} & 0.0011 & \textbf{0.0069} & \textbf{0.5592} & \bf{0.0002}\\ \hline
        $\textit{MD}_\text{dif}$ & \textbf{0.0797} & 0.0007 & \bf{0.0041} & \textbf{0.0470} & \textbf{0.0074} & 0.0051 & \bf{0.0003} & \textbf{0.3638} & \textbf{0.0002}\\ \hline
        $\textit{MD}_\text{dif,col}$ & \bf{0.0513} & \bf{0.0004} & \bf{0.0041} &\bf{0.0620} & 0.1140 & \bf{0.0005} & \bf{0.0013} & \textbf{0.4242} & \textbf{0.0006}\\ \hline
    \end{tabular}
    }
\end{table*}

\subsection{Evaluation: Molecular Generative Efficiency}
\label{subsec:eff_eval}
The training loss and reconstruction success rate can be important indicators for molecular generation verification. 
However, they are insufficient to judge of a performance of molecular generative models. 
In this subsection, our proposed methods are evaluated in terms of molecular generative efficiency. 
The generative efficiency is the rate of the generated molecule satisfying validity, uniqueness, and novelty. 
The validity means that the generated molecule has a sound structure determined using the RDKIT package \cite{rdkit}. 
A molecule is said to be novel if it is not in the train database. 
The uniqueness refers to how many molecules are generated without duplication.
For example, if the generative efficiency is 0.9, 90 molecules are sound, distinct and novel when the molecular generation is tried 100 times. 

Instead of approximated latent space from a given molecule, latent variables are sampled and new molecules are generated as in Eq.~\eqref{eq:infer}. 
%Of course, the encoder of VAE is not used for this experiment.
In order to generate molecules, it is necessary to specify the condition that generated molecules should have.
Each molecule of ZINC dataset has three properties: molWt, LogP and QED.
One of the properties were determined manually as below, and the other properties were sampled from the conditional probability distribution.

We specified two types of conditions: in-domain and y-extrapolation (OOD).
In the in-domain case, the mean $\mu$ and lower and upper limits of 90\% confidence interval ($\mu \pm 1.645\sigma$) of train dataset were used (molWt $\in \{330, 434, 230\}$, LogP $\in\{2.457, 4.816, 0.098\}$ , QED $\in \{0.7318, 0.9598, 0.5038\}$). 
Meanwhile, in the OOD case, outlier values ($\mu \pm 4\sigma$) were used (molWt $\in \{580, 84\}$ , LogP $\in \{8.194, -3.281\}$ , QED $\in \{0.1775, 1.2861\}$ ).
%These values do not even exist in the training dataset as shown in Figure~\ref{fig:distrib}.
For each condition, 2,000 molecules were generated by each method. (For each model, 30,000-time generation was tried.)
 
Table~\ref{tab:eff} illustrates generative efficiency rates for each property. 
In the in-domain case, \textit{MD}s except \textit{MD}$_\text{col}$ for QED showed higher efficiency than \textit{ControlVAE}, and \textit{MD} outperformed 3.5\% relative to \textit{ControlVAE} . 
In the OOD case, it showed higher relative improvement than when the in-domain case (In-domain and OOD cases' relative improvement of \textit{MD}$_\text{dif,col}$: 2.2\%, 31.4\%).

\begin{figure}
  \centering
    \begin{tabular}{ccc}
	%\includegraphics[width=0.18\textwidth]{mol1-1.PNG} &
	%\includegraphics[width=0.06\textwidth]{mol1-2.PNG} \\
	%molwt: $580\rightarrow580.08$ & 
	%molwt: $84\rightarrow84.07$\\
	%\includegraphics[width=0.12\textwidth]{mol1-3.PNG} &
	%\includegraphics[width=0.18\textwidth]{mol1-4.PNG} \\
	%LogP: $8.194\rightarrow8.1932$ & 
	%LogP: $-3.281\rightarrow-3.2816$ \\
	%\includegraphics[width=0.18\textwidth]{mol1-5.PNG} &
	%\includegraphics[width=0.18\textwidth]{mol1-6.PNG} \\
	%QED: $0.9598\rightarrow0.9481$ & 
	%QED: $0.1775\rightarrow0.1777$\\
	\includegraphics[width=0.16\textwidth]{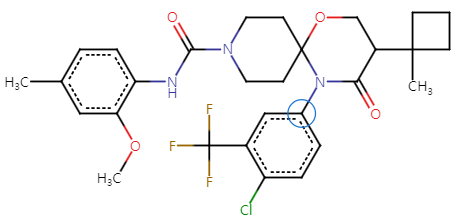} &
	\includegraphics[width=0.05\textwidth]{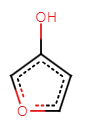} \\
	molwt: $580\rightarrow580.047$ & 
	molwt: $84\rightarrow84.074$\\
	\includegraphics[width=0.10\textwidth]{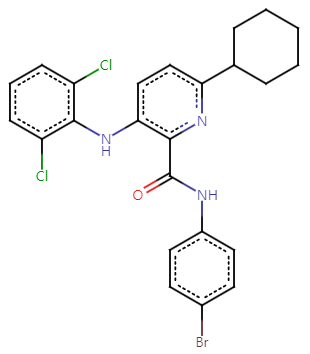} &
	\includegraphics[width=0.05\textwidth]{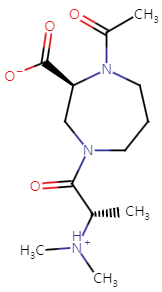} \\
	LogP: $8.194\rightarrow8.1945$ & 
	LogP: $-3.281\rightarrow-3.2813$ \\
	\includegraphics[width=0.10\textwidth]{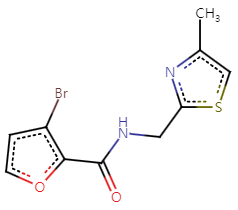} &
	\includegraphics[width=0.10\textwidth]{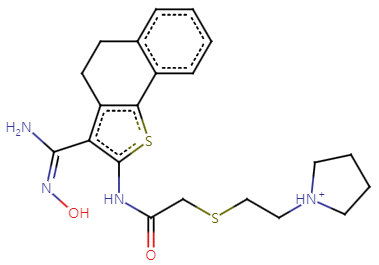} \\
	QED: $0.9598\rightarrow0.948094$ & 
	QED: $0.1775\rightarrow0.177306$\\
  \end{tabular}
  \caption{Top1 generated molecules of the proposed method.The script below each molecular structure is the order of properties, conditions, and predicted values by RDKIT.}
  %\caption{Top1 generated molecules of \textif{MD}s with OOD's conditions: The script below each molecular structure is the order of properties, conditions, and predicted values by RDKIT (QED's condition 0.9598 is also OOD case.).}
  \label{fig:Top1Mol}
\end{figure}

\subsection{Evaluation: Conditional Satisfaction}
\label{subsec:cs}
Other important indicator is a conditional satisfaction for novel molecules. 
In the actual molecular generative task, the discovery of new molecules with properties close to the target properties is essential. %, and the number of novel molecules is not an essential factor. 
For this reason, we measured a mean absolute error (MAE) of the \textit{top 1} molecule having the smallest MAE to the input conditions, as shown in Table~\ref{tab:satis}. 
Three properties of generated molecules were all calculated using RDKIT. 
In the table, the dash (-) means that the model did not generate even a single valid molecule. 
When the condition is $\text{QED}=1.2861$, most models showed high MAE because QED=1.2861 is a value that could not exist physically in QED property (The maximum QED value from RDKIT is 0.948.).
In the case of In-domain, most proposed models achieved a lower MAE value than \textit{ControlVAE}. 
In the case of OOD, \textit{MD}$_\text{diff,col}$ outperformed \textit{ControlVAE} except the condition $\text{MolWt}=84$. In particular, \textit{MD}s showed a sensational performance when the LogP's condition was -3.281. In a nutshell, the proposed methods averagely showed a lower MAE than \textit{ControlVAE} and \textit{Base}.
 Top 1 molecules of the proposed methods at the OOD case are presented in Figure~\ref{fig:Top1Mol}.

\subsection{Evaluation: Performance Comparison as the Number of Decoder}
\label{subsec:num}
Until now, we showed our proposed method's performance when the number of decoder was 3. In this subsection, we evaluate \textit{MD}$_\text{diff,col}$'s performance according to the number of decoders. Since \textit{Base} showed a poor performance, we compared \textit{ControlVAE} with \textit{MD}$_\text{diff,col}$. Of course, \textit{ControlVAE} was trained as several model sizes. Table~\ref{fig:MD} (b) showed a reconstruction loss as the number of decoders and model size (mb). Each model trained until 100 epoch, and the number of decoders of \textit{MD}$_\text{diff,col}$ is from 2 to 7. The size of encoder of \textit{MD}$_\text{diff,col}$ is the same. In all cases, \textit{MD}$_\text{diff,col}$ showed a lower reconstruction loss than \textit{ControlVAE}. 

%Table~\ref{fig:MD} (c) showed a reconstruction success rate as model sizes. the proposed method outperformed $\textit{ControlVAE}$ at the not only unseen case but also seen case.  

Table~\ref{fig:MD} (c) showed a reconstruction success rate as model sizes. $\textit{MD}_\text{diff,col}$ outperformed $\textit{ControlVAE}$ at the not only unseen case but also seen case. In particular, although $\textit{ControlVAE}$ could not maintain its OOD performance as the model size, $\textit{MD}_\text{diff,col}$ showed a stable performance.

\section{Conclusion}
In this paper, we propose a simple method % MD-VAE 
for molecular generation task. Our main idea is to  maintain multiple decoders while sharing a single encoder.
%, i.e., it is a type of ensemble techniques.
To facilitate synergy between decoders, we introduce a collaborative loss function. We also propose to utilize a different latent variable for each decoder in order to diversify decoders while they collaborate with others to achieve a common goal. 
%Experimental results showed that MD-VAE outperforms VAE with k-annealing and controllable VAE in terms of various measures.
%Experimental results showed that MD-VAE incorporating both the collaborative loss and different latent variables outperforms VAE with k-annealing and controllable VAE in terms of various measures: training loss, reconstruction success rate, molecular generative efficiency and conditional satisfaction (Top1 molecule).
In our experiments, especially for out-of-distribution conditions, MD-VAE is far better than baselines %controllable VAE 
with respect to efficiency and conditional satisfaction of generated molecules.

\bibliographystyle{named}
\bibliography{ijcai22.bib}

\end{document}